\documentclass[lettersize,journal]{IEEEtran}
\usepackage{amsmath,amsfonts}
\usepackage{algorithmic}
\usepackage{algorithm}
\usepackage{array}
\usepackage[caption=false,font=normalsize,labelfont=sf,textfont=sf]{subfig}
\usepackage{textcomp}
\usepackage{stfloats}
\usepackage{url}
\usepackage{verbatim}
\usepackage{graphicx}
\usepackage{cite}
\usepackage{booktabs}
\usepackage{color}
\usepackage{amssymb}
\usepackage{bm}
\usepackage{amsmath}
\usepackage{xcolor}
\usepackage{multicol}
\usepackage{multirow}
\usepackage{colortbl}

\definecolor{cvprblue}{rgb}{0.21,0.49,0.74}
\usepackage[pagebackref,breaklinks,colorlinks]{hyperref}

\begin{document}

\title{Multi-Resolution Alignment for Voxel Sparsity in\\Camera-Based 3D Semantic Scene Completion}

\author{Zhiwen Yang, and Yuxin Peng,~\IEEEmembership{Fellow,~IEEE,}
\thanks{Date of current version 31 December 2025. This work was supported by the grants from the National Natural Science Foundation of China (62525201, 62132001, 62432001) and Beijing Natural Science Foundation (L247006, L257005). \textit{(Corresponding author: Yuxin Peng.)}}
\thanks{Zhiwen Yang, and Yuxin Peng are with the Wangxuan Institute of Computer Technology, Peking University, Beijing 100871, China (e-mail: pengyuxin@pku.edu.cn).}}

\markboth{IEEE TRANSACTIONS ON IMAGE PROCESSING}%
{Shell \MakeLowercase{\textit{et al.}}: A Sample Article Using IEEEtran.cls for IEEE Journals}


\maketitle

\begin{abstract}
Camera-based 3D semantic scene completion (SSC) offers a cost-effective solution for assessing the geometric occupancy and semantic labels of each voxel in the surrounding 3D scene with image inputs, providing a voxel-level scene perception foundation for the perception-prediction-planning autonomous driving systems. Although significant progress has been made in existing methods, their optimization rely solely on the supervision from voxel labels and face the challenge of voxel sparsity as a large portion of voxels in autonomous driving scenarios are empty, which limits both optimization efficiency and model performance. To address this issue, we propose a \textit{Multi-Resolution Alignment (MRA)} approach to mitigate voxel sparsity in camera-based 3D semantic scene completion, which exploits the scene and instance level alignment across multi-resolution 3D features as auxiliary supervision. 
Specifically, we first propose the Multi-resolution View Transformer module, which projects 2D image features into multi-resolution 3D features and aligns them at the scene level through fusing discriminative seed features.
Furthermore, we design the Cubic Semantic Anisotropy module to identify the instance-level semantic significance of each voxel, accounting for the semantic differences of a specific voxel against its neighboring voxels within a cubic area. 
Finally, we devise a Critical Distribution Alignment module, which selects critical voxels as instance-level anchors with the guidance of cubic semantic anisotropy, and applies a circulated loss for auxiliary supervision on the critical feature distribution consistency across different resolutions.
Extensive experiments on the SemanticKITTI and SSCBench-KITTI-360 datasets demonstrate that our MRA approach significantly outperforms existing state-of-the-art methods, showcasing its effectiveness in mitigating the impact of sparse voxel labels.
The code is available at \textcolor{blue}{\href{https://github.com/PKU-ICST-MIPL/MRA_TIP}{https://github.com/PKU-ICST-MIPL/MRA\_TIP}}.
\end{abstract}



\begin{figure*}[t]
\centering
\includegraphics[width=1.0\linewidth]{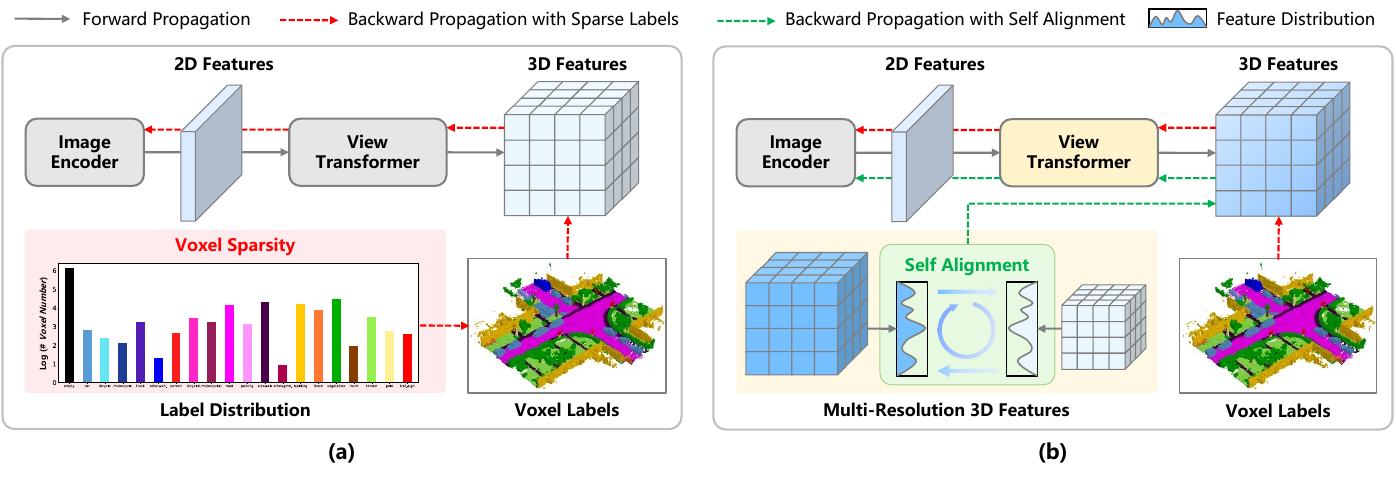} 
\caption{
(a) Supervision signals from voxel labels suffer from the challenge of voxel sparsity.
(b) Our self alignment across multi-resolution 3D feature distributions serve as auxiliary supervision signals for addressing the voxel sparsity.
}
\label{motivation}
\end{figure*}

\section{Introduction}
\IEEEPARstart{A}{ccurate} perception of the surrounding 3D scenes acts as the cornerstone of the perception-prediction-planning autonomous driving systems and facilitates multiple downstream tasks such as object detection, trajectory navigation and motion planning. In recent years, 3D semantic scene completion (SSC) has garnered increased attention for its ability to generate voxel-level scene understanding by predicting both the geometric occupancy and semantic labels of each voxle in the entire 3D scene. Due to the high complexity of outdoor autonomous driving scenarios, LiDAR points have been the primary input modality~\cite{roldao2020lmscnet},~\cite{cheng2021s3cnet},~\cite{rist2021semantic},~\cite{an2024multimodality} since point cloud data maintains the structural information of the surrounding 3D scenes. Despite the effectiveness of LiDAR-based methods~\cite{yan2021sparse},~\cite{xia2023scpnet},~\cite{li2023lode},~\cite{mei2023ssc},~\cite{an2024rethinking} on 3D semantic scene completion, they suffer from the expensive cost and limited scalability of LiDAR sensors, restricting their extensive applications in real-world scenarios.

Recently, camera-based SSC methods have emerged as a more cost-effective and scalable alternative for 3D semantic scene completion with only 2D image inputs. The key challenges for camera-based SSC methods arise from the dimension gap between 2D inputs and 3D outputs, making it difficult to generate accurate 3D representations. MonoScene~\cite{cao2022monoscene} first proposes to lift 2D image features into 3D volumes with depth-based mapping projection, followed by a 3D U-net processing the lifted 3D features. Then, the classic pipeline consisting of the image encoder, view transformer, and voxel model has been widely adopted in camera-based SSC methods~\cite{miao2023occdepth},~\cite{wei2023surroundocc},~\cite{zhang2023occformer},~\cite{li2023voxformer}. Subsequent efforts have been made to improve the procession of 3D features, such as exploiting local structures of objects and spatial differences from different perspectives~\cite{xue2024bi}, integrating instance semantics and scene contexts~\cite{jiang2024symphonize}, and designing dense-sparse-dense structures for effective and efficient voxel aggregation~\cite{mei2024camera}.

Although existing camera-based SSC methods have made significant progress with carefully designed voxel models to generate improved 3D features, they rely solely on the sparse 3D voxel labels as the primary supervision signal for model training and optimization, struggling with the challenge of \textbf{voxel sparsity}. As illustrated in Fig~\ref{motivation} (a), we present the voxel-level label distributions over the SemanticKITTI~\cite{behley2019semantickitti} validation dataset, where the y-axis represents the logarithm of voxel numbers and the x-axis between empty voxels (represented in black) and occupied voxels (represented in other colors).  
The large proportion (over $92.9\%$) of empty voxels poses significant challenges for model optimization: (1) the sparse and uneven distribution of supervision signals leads to limited gradient propagation in occupied regions, resulting in sub-optimal feature learning and slower convergence; (2) the loss function becomes dominated by the abundance of empty voxels, biasing the model towards minimizing trivial errors on non-informative regions instead of focusing on semantically meaningful structures.

To enable more effective training and optimization process, we propose a \textbf{M}ulti-\textbf{R}esolution \textbf{A}lignment (\textbf{MRA}) approach for addressing the voxel sparsity, which performs the view transformation process through the fusion of multi-resolution 3D features and takes advantage of the self alignment of feature distributions across different resolution levels.
Compared to previous methods~\cite{li2023voxformer, mei2024camera, wang2024not}, our core novelty lies in the design of a multi-resolution alignment framework specifically addressing voxel sparsity in SSC, proposing a sparsity-aware alignment mechanism to systematically compensate for sparse voxel supervision, a challenge that existing SSC methods do not explicitly target.
As shown in Fig~\ref{motivation} (b), the circulated feature distribution alignment serves as an auxiliary supervision signal for the backward propagation, alleviating the voxel sparsity in the backward propagation with sparse labels.
By encouraging consistent feature representations across different resolution levels, our approach compensates for the sparsity of voxel labels and enhances the learning of discriminative semantics in critical regions. This alignment-driven supervision ensures more effective gradient flow, alleviates the negative impact of voxel sparsity, and enables robust and efficient semantic scene completion.
Specifically, we first propose the Multi-resolution View Transformer (MVT) module to project the 2D image features into multi-resolution 3D features. The multi-resolution projection block first projects the 2D image features into pre-defined multi-resolution voxel grids with intrinsic and extrinsic camera parameters. The projected coarse 3D features are then fed into the seed feature alignment block, where multi-resolution discriminative semantics are fused and propagated to the entire 3D scene.
Additionally, we design the Cubic Semantic Anisotropy (CSA) module to represent the semantic-aware significance of each voxel, accounting for different semantic-aware structural information. Compared with the Local Geometric Anisotropy (LGA)~\cite{li2019depth} for in-door scenarios, we have made two improvements with CSA for autonomous driving scenes. Firstly, due to the inherent complexity of multiple semantic categories in autonomous driving scenes, we execute the semantic reassignment to cluster similar semantic classes, like ``\textit{bicyclist}" and ``\textit{motorcyclist}", facilitating more efficient and reasonable computation of semantic differences. Secondly, we calculate the cubic anisotropy with a comprehensive aggregation of the surface-, edge-, and vertex-adjacent semantic differences with in a cubic neighboring area, which effectively distinguish between voxels along object boundaries and inside the object.
Finally, we introduce the Critical Distribution Alignment (CDA) module to exploit the consistency between multi-resolution feature distributions as auxiliary supervision signals. To begin with, we first employ the critical voxel selection block, which selects critical voxels under the guidance of occupancy confidence and cubic semantic anisotropy. The critical voxels act as instance-level anchors for the 3D scene, indicating discriminative semantics at significant voxels. Therefore, we further design the circulated loss to promote the critical feature distribution consistency across different resolution levels, providing auxiliary supervision signals for consistent multi-resolution 3D features.
Extensive experiments and analyses on the SemanticKITTI and SSCBench-KITTI-360 datasets validate the effectiveness of our MRA approach.

We summarize the contributions of this paper as follows:
\begin{itemize}
    \item We identify the challenge of voxel sparsity in camera-based 3D semantic scene completion, and introduce the Multi-Resolution Alignment (MRA) approach, which is designed with a sparsity-aware multi-resolution alignment architecture, forming a top-down pipeline from identifying contextual semantic significance to aligning global feature consistency.
    \item Multi-resolution view transformer is proposed to project 2D image features into multi-resolution 3D features with intrinsic and extrinsic camera parameters, and exploits scene-level alignment with seed feature fusion and discriminative semantic propagation.
    \item Cubic semantic anisotropy is designed to identify the semantic-aware significance of each voxel, which clusters similar classes with semantic reassignment and accounts for the voxel-wise semantic differences within a $3\times 3\times 3$ cubic neighboring area.
    \item Critical distribution alignment takes advantage of occupancy confidence and cubic semantic anisotropy to select critical voxels as instance-level anchors, and then adopts the circulated loss as auxiliary supervision signals on the critical distribution consistency, effectively addressing voxel sparsity in label supervision.
\end{itemize}

The rest of this paper is organized as follows: Section \ref{sec:relatedwork} provides a brief review of related work, including both LiDAR-based and camera-based SSC methods. Section \ref{sec:approach} presents a detailed description of the proposed Multi-Resolution Alignment (MRA) approach. Section \ref{sec:experiment} discusses the experimental results, including quantitative analyses and visualization results. Finally, Section \ref{sec:conclusion} concludes this paper and outlines directions for future work.

\section{Related Works}
\label{sec:relatedwork}

In this section, we briefly review the relevant methods and literature for the 3D semantic scene completion (SSC) task, whose objective is to predict both the occupancy and semantic labels for each voxel in the targeted 3D scene, first introduced by SSCNet~\cite{song2017semantic}. Based on different input modalities, SSC methods can be broadly divided into two main streams: LiDAR-based and camera-based SSC methods.

\subsection{LiDAR-based SSC Methods.} 
LiDAR-based methods have long been the dominant solutions to the SSC task due to the 3D structural information contained in the LiDAR point clouds. UDNet~\cite{zou2021up} first leverages a 3D U-Net architecture for generating scene predictions directly from the voxel grids constructed by LiDAR point clouds, but suffers from high computational overhead due to large amount of voxel volumes in the scene. To improve computational efficiency, LMSCNet~\cite{roldao2020lmscnet} employs 2D convolutional networks to reduce computational cost with lightweight encoding of voxel features. In a similar vein, SGCNet~\cite{zhang2018efficient} adopts spatial group convolutions to split input voxels into distinct groups, enabling sparse 3D convolutions for improved computational efficiency. Further advancements in LiDAR-based SSC methods have focused on improving 3D scene representations by exploring multi-view fusion for the completion of sparse scene representations~\cite{cheng2021s3cnet}, as well as designing local implicit functions as continuous scene representations~\cite{rist2021semantic}. Another promising direction is to take advantage of knowledge distillation, as demonstrated by~\cite{xia2023scpnet}, which transfers rich multi-frame information to single-frame models with lightweight architectures. Recently, there has been increasing interest in integrating semantic segmentation with scene completion tasks for improved accuracy and robustness. QGPA~\cite{he2023prototype} targets at the challenge of large intra-class feature variance in 3D data, and proposes to adapt class prototypes from support to query feature spaces for few-shot learning, further projecting category word embeddings to generate prototypes for zero-shot scenarios. JS3C-Net~\cite{yan2021sparse} introduces a point-voxel interaction module to facilitate knowledge fusion between semantic segmentation and scene completion tasks, while SSA-SC~\cite{yangsemantic} proposes a double branch network that assists scene completion with semantic information from the semantic segmentation branch. Similarly, SSC-RS~\cite{mei2023ssc} combines multiple branches to hierarchically fuse semantic and geometric features for improved semantic scene completion performance. Despite achieving considerable progress, LiDAR-based SSC methods still struggle with the challenge of high computational cost due to the extensive amount of LiDAR points, restricting their scalability to larger real-world scenarios.

\subsection{Camera-based SSC Methods}
In recent years, camera-based SSC methods~\cite{miao2023occdepth, yao2023ndc, tong2023scene, wang2024h2gformer, li2019depth, wang2024not} have attracted increasing attention for their cost-effectiveness and easy deployment with only RGB image inputs in autonomous driving applications. One of the first fully visual SSC framework, MonoScene~\cite{cao2022monoscene}, introduces an innovative method, which samples image features along lines of sight into voxel grids and utilizes the 3D U-net architecture for voxel occupancy predictions. Other methods adopt the Bird's-Eye-View (BEV) representation as a top-down integration of image features for 3D scene representations. The pioneering work LSS~\cite{philion2020lift} lifts input images into frustums of features and then splats all frustums into rasterized BEV grids. Such paradigm is further enhanced by BEVDet~\cite{huang2021bevdet}, which introduces a general BEV-based pipeline to align multi-frame image features for improved 3D scene comprehension. Subsequent efforts have been made to refine the BEV representations, incorporating techniques such as inverse perspective mapping~\cite{hu2021fiery}, camera-aware depth estimation and refinement~\cite{li2023bevdepth}, dynamic temporal stereo information integration~\cite{li2023bevstereo}, angle and radius aware rasterization~\cite{liu2023vision}, and cross-attention layers with BEV queries~\cite{li2022bevformer}. In addition, TPVFormer~\cite{huang2023tri} proposes the Tri-Perspective View (TPV) representation, which decomposes voxel grids onto three orthogonal view planes for more efficient and comprehensive scene encoding. Voxformer~\cite{li2023voxformer} introduces a two-stage framework that combines class-agnostic query proposals and class-specific semantic segmentation, effectively diffusing discriminative semantics from seed voxels to the whole scene. OccFormer~\cite{zhang2023occformer} designs a dual-path transformer network to improve scene completion through the mask-wise prediction paradigm. Bi-SSC~\cite{xue2024bi} tackles occlusion issues by leveraging the neighboring structures of target objects and spatial differences observed from different viewpoints. Symphonize~\cite{jiang2024symphonize} devises a scene-from-instance paradigm to facilitate intricate interactions between the image and volumetric domains. SGN~\cite{mei2024camera} further proposes a one-stage SSC framework with the dense-sparse-dense design, ensuring clear segmentation boundaries for more accurate 3D semantic scene completion.\par

Although significant improvements have been demonstrated, the above camera-based 3D semantic scene completion methods rely solely on the annotated voxel labels for training and optimization, which suffer from the inherent voxel sparsity in autonomous drive scenes, limiting the model's scene completion performance with disturbance from large proportion of empty voxels. To address this issue, we propose the Multi-Resolution Alignment (MRA) approach to exploit the scene-level and instance-level alignment between multi-resolution 3D features, which provide auxiliary supervision signals for training and optimization. Therefore, the voxel sparsity is alleviated with complementary supervisions from annotated voxel labels and multi-resolution feature alignment, leading to improved 3D semantic scene completion performance.

\begin{figure*}[!t]
  \centering
  \includegraphics[width=1.0\linewidth]{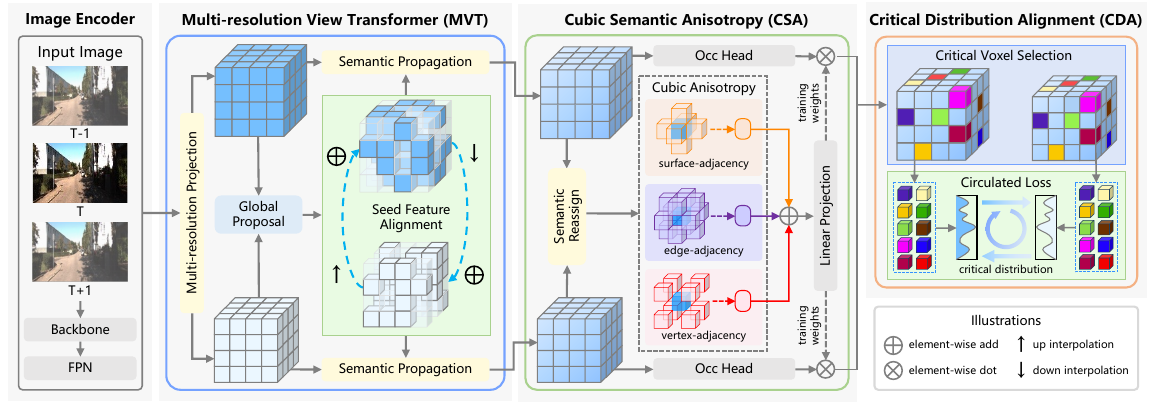}
  \caption{The overall architecture of our MRA framework. The Multi-resolution View Transformer (MVT) module projects 2D image features into multi-resolution 3D features and conduct seed feature alignment for the propagation of discriminative semantics to the whole scene. The Cubic Semantic Anisotropy (CSA) module identifies the semantic-ware significance of each voxel with semantic reassignment and aggregation of cubic semantic differences. The Critical Distribution Alignment (CDA) module selects critical voxels with the guidance of CSA and employs the circulated loss as auxiliary supervision on the critical distribution consistency.}
  \label{approach}
\end{figure*}

\section{Approach}
\label{sec:approach}

\subsection{Overview}
Figure~\ref{approach} illustrates the overall framework of our proposed Multi-Resolution Alignment (MRA) approach, which comprises four key components: (1) an Image Encoder for extracting 2D image features; (2) a Multi-resolution View Transformer (MVT) module that projects 2D image features into multi-resolution 3D features and aligns discriminative seed features to propagate semantics throughout the entire scene; (3) a Cubic Semantic Anisotropy (CSA) module that evaluates the semantic-aware significance of each voxel with semantic reassignment and aggregation of semantic differences within a cubic neighboring area; (4) a Critical Distribution Alignment (CDA) module that identifies critical voxels based on the guidance from CSA and adopts the circulated loss as auxiliary supervision to promote the feature distribution consistency across critical voxels of different resolutions.
MRA is designed as a multi-resolution alignment framework specifically addressing voxel sparsity, with the above modules forming a top-down alignment pipeline, where MVT lays the geometric foundation, CSA focuses the alignment scope, and CDA refines the feature consistency.

\subsubsection{Problem Setup.} The target of SSC is to predict the geometry and semantics of 3D scenes in front with monocular~(left or right) input images $I^{\rm rgb}$. The output is represented as a voxel grid $Y\in\mathbb{R}^{X\times Y\times Z}$, where $X, Y, Z$ correspond to the grid's resolution. Each voxel is categorized as either empty denoted by $c_0$ or occupied by one of the semantic classes in $C\in\{c_1,\cdots,c_N\}$, where $N$ represents the number of semantic classes. In essence, SSC aims to train a model $Y=\Theta(I^{\rm rgb})$ that can generate a 3D semantic prediction $Y$ closely matching the ground truth $\Bar{Y}$.
\subsubsection{Image Encoder.} We employ the ResNet-50~\cite{he2016deep} network with FPN~\cite{lin2017feature} to build the image encoder for extracting 2D features $F^{\rm 2D}\in\mathbb{R}^{N_t\times C\times H\times W}$ from input RGB images, where $N_t$ represents the number of temporal image inputs, $C$ denotes the number of feature channels, and $(H,W)$ specifies the resolution of extracted image features.

\subsection{Multi-resolution View Transformer}
The Multi-resolution View Transformer (MVT) module consists of two blocks to project 2D image features into multi-resolution 3D features. The Multi-resolution Projection block first projects 2D image features into pre-defined multi-resolution voxel grids with intrinsic and extrinsic camera parameters. Then the Seed Feature Alignment block further refines the coarse 3D semantics, which aligns multi-resolution seed features filtered by depth-aware proposals and propagates discriminative semantics to the whole scene.

\subsubsection{Multi-resolution Projection}
Following~\cite{cao2022monoscene, mei2024camera}, we first project 2D image features into coarse multi-resolution 3D features with the 2D-3D projection functions based on camera parameters. Specifically, for each voxel grid at resolution level $i$ with the size of $(X_i, Y_i, Z_i)$, we denote the centroid coordinates of total $X_i\times Y_i\times Z_i$ voxels as $\bm{x_i}\in\mathbb{R}^{X_i\times Y_i\times Z_i\times 3}$, formulating the projection correlation $\pi(\bm{x_i})$ with the intrinsic camera parameter matrices $\bm{K}$ and extrinsic camera parameter matrices $\bm{T}=[\bm{R}, \bm{t}]$ directly provided in the KITTI~\cite{geiger2012we} dataset. For a specific point $p_i=(x_p^i, y_p^i, z_p^i)$ in $\bm{x_i}$, the projection correlation indicates the correspondence with respect to any image pixel $(u, v)$ as follows:
\begin{align}
    [x_p^i, y_p^i, z_p^i]^{T} &= \bm{R}\cdot p_i + \bm{t} \; , \\
    z_p^i \circ [u, v, 1]^{T} &= \bm{K}\cdot [x_p^i, y_p^i, z_p^i]^{T} \; ,
\end{align}
where $\circ$ denotes the element-wise product. Since one voxel may sample features from multiple images out of the temporal input images, we compute the average feature for voxels within the field of view (FOV) and assign zero to those outside the field of view:
\begin{equation}
    \bm{F}_{i}^{\rm 3D} = W_i \cdot \sum\limits_{t=1}^{N_t} [\Phi_{\pi(\bm{x_i})}(\bm{F}_{t}^{\rm 2D}) \cdot \bm{M}_{t}^{\rm FOV}] \; ,
\end{equation}
where $\bm{F}_{i}^{\rm 3D}\in\mathbb{R}^{C\times X_i\times Y_i\times Z_i}$ is the sampled coarse 3D features of resolution level $i$, $\bm{F}_{t}^{\rm 2D}$ is the 2D features of the $t$-th temporal image $\bm{I}_t$, $\Phi_{\pi(\bm{x_i})}(\cdot)$ is the feature sampling function based on projection correlation $\pi(\bm{x_i})$, and $\bm{M}_{t}^{\rm FOV}$ is the binary mask representing the field of view of $\bm{I}_{t}$. $W_i$ is the weight matrix defined as follows:
\begin{equation}
{W_{i}^{p}} = \begin{cases}
1 / \delta_{p_i}, &\delta_{p_i} > 0 \; , \\ 
1, & \delta_{p_i} = 0 \; ,
\end{cases}
\end{equation}
where $\delta_{p_i}$ is the total number of input images that perceive the view of point $p_i$.

\subsubsection{Seed Feature Alignment}
To further refine the projected coarse 3D features, we select seed features with global voxel proposals to align the multi-resolution 3D features and propagate fine-grained semantics to the whole scene. Following~\cite{li2023voxformer} and~\cite{mei2024camera}, we employ the pre-trained Mobliestereonet~\cite{shamsafar2022mobilestereonet} to predict the depth map of input images and then back-projects the depth map into 3D point clouds with intrinsic and extrinsic camera parameters $(\bm{K}, \bm{T})$:
\begin{equation}
    p = \bm{R}^{-1} \cdot [\bm{K}^{-1}\cdot (d_{(u,v)}\circ [u, v, 1]^{T}) - \bm{t}] \; ,
\end{equation}
where $p$ denotes the coordinates of the projected 3D points, and $d_{(u, v)}$ is the predicted depth value at pixel $(u, v)$. Then the projected point clouds $\bm{P}$ are fed into a multi-scale heuristic head to generate class-agnostic occupancy proposals of different resolutions $\{\bm{O}_{i}\in\mathbb{R}^{X_i\times Y_i\times Z_i}\}$, and the seed features are filtered with a confidence threshold $\theta$:
\begin{equation}
    \bm{F}_{i}^{\rm seed} = \bm{F}_{i}^{\rm 3D}[:, \bm{O}_i>\theta] \; ,
\end{equation}
where $\bm{F}_{i}^{\rm seed}$ is the selected seed features with discriminative semantics of resolution level $i$. The multi-resolution seed features are then aligned mutually with a seed alignment layer consisting of a trilinear interpolation layer for upsampling and an average pooling layer for downsampling:
\begin{align}
    \bm{F}_{1}^{\rm seed} &= \bm{F}_{1}^{\rm seed} + {\rm Up}(\bm{F}_{2}^{\rm seed}) \; , \\
    \bm{F}_{2}^{\rm seed} &= \bm{F}_{2}^{\rm seed} + {\rm Down}(\bm{F}_{1}^{\rm seed}) \; ,
\end{align}
where ${\rm UP}(\cdot), {\rm Down}(\cdot)$ indicate the upsampling and downsampling operations respectively. Based on the fused seed features, we further execute semantic propagation with the ASPP~\cite{chen2017deeplab} layer, diffusing discriminative semantics from seed voxels to the entire 3D scene.

\begin{figure}[!t]
\centering
\includegraphics[width=1.0\columnwidth]{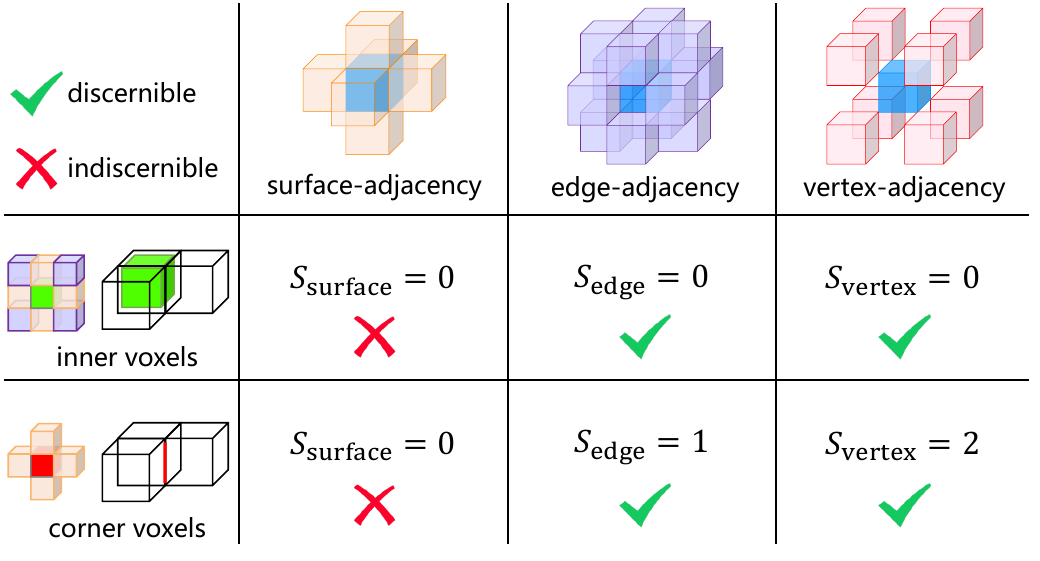}
\caption{Illustrations of the cubic semantic anisotropy for inner voxels inside the object and corner voxels on the object boundary. A thorough consideration of the surface-, edge-, and vertex-adjacency within the cubic neighborhood is crucial for accurately identifying voxel-wise semantic-aware significance.}
\label{sga}
\end{figure}

\subsection{Cubic Semantic Anisotropy}
To tackle the challenge of voxel sparsity and enable effective self-alignment across multi-resolution 3D features, it is crucial to identify critical voxels with discriminative semantics that can possess high semantic consistency with neighboring voxels and effectively reflect the semantic context of surrounding areas, serving as representatives for neighboring regions.
While the Local Geometric Anisotropy (LGA)~\cite{li2019depth} method has been introduces to describe the position-aware significance of each voxel concerning the semantic differences against the six surface-adjacent voxels, it targets at in-door scenarios and suffers from two drawbacks with respect to out-door autonomous driving scenes:
\begin{itemize}
    \item The out-door autonomous driving scenes are inherently more complex and feature a greater variety of semantic categories compared to in-door scenarios, including highly similar objects like ``\textit{bicycle}" and ``\textit{motorcycle}" as well as distinct objects such as ``\textit{vegetation}" and ``\textit{building}". However, LGA overlooks the varying relationships among target semantic categories and treats all objects equally. Thus, we propose the Semantic Reassignment block to cluster similar semantic categories for more effective and reasonable identification of semantic differences.
    \item LGA only takes into account the semantic differences between each voxel and its six ``surface-adjacent' voxels, overlooking the ``edge-adjacent" and ``vertex-adjacent" voxels, which are crucial for distinguishing between voxels located along the object boundaries and those inside the object, as illustrated in Fig~\ref{sga}. Therefore, we propose the Cubic Anisotropy block for a thorough aggregation of semantic differences within the cubic neighborhood and accurate identification of semantic-aware significance for each voxel.
\end{itemize}
Detailed descriptions on the Semantic Reassignment block and Cubic Anisotropy block are presented as follows:

\subsubsection{Semantic Reassignment}
In consideration of the inherent complexity and diversity of semantic categories, we cluster similar semantic categories for effective and reasonable identification of semantic differences across distinct categories. Firstly, all categories are divided into two groups $\mathbb{C}_{1}=\{``foreground", ``background"\}$, where ``\textit{foreground}" category contains the dynamic, small-scale objects and ``\textit{background}" category includes the static, large-scale objects. Furthermore, we refine the above division according to the semantic similarities across the target categories, resulting in the following refined separation:
\begin{itemize}
    \item ``\textit{vehicle}": car, bicycle, motorcycle, truck, other-veh.
    \item ``\textit{human}": person, bicyclist, motorcyclist
    \item ``\textit{ground}": road, parking, sidewalk, other-grnd., terrain
    \item ``\textit{building}": building
    \item ``\textit{infrastructure}": fence, pole, traf.-sign
    \item ``\textit{plant}": vegetation, trunk
\end{itemize}
It is worth noting that our semantic reassignment scheme is designed as a coarse-grained grouping based on intrinsic semantic similarities, which is intended to be task-agnostic and generalizable across datasets.
Based on the above semantic groups, we reassign the voxels belonging to the same group with a unified semantic label:
\begin{equation}
    \bm{V}_{\rm re} = [{\rm idx}|\bm{V}_{\rm gt}\in \mathbb{C}_2({\rm idx})] \; ,
\end{equation}
where $\bm{V}_{\rm gt}\in\mathbb{R}^{X\times Y\times Z}$ is the ground truth voxel labels, $\mathbb{C}_2$=\{``\textit{vehicle}", ``\textit{human}",``\textit{ground}", ``\textit{building}", ``\textit{infrastructure}", ``\textit{plant}"\} denotes the refined groups of all semantic categories, and $\bm{V}_{\rm re}\in\mathbb{R}^{X\times Y\times Z}$ is the reassigned semantic labels for the following computation.

\begin{table*}
  \centering
  \caption{Semantic scene completion comparison results against the state-of-the-art camera-based methods on the SemanticKITTI~\cite{behley2019semantickitti} validation set. Best results are highlighted in \textbf{bold}, and second-best results are \underline{underlined}.}
  \resizebox{1.0\linewidth}{!}{
  \begin{tabular}{@{}l|cccccc@{}}
    \toprule
    \textbf{Method} & {\small MRA~(Ours)} & {\small SGN~\cite{mei2024camera}} & {\small HASSC\cite{wang2024not}} & {\small VoxFormer~\cite{li2023voxformer}} & {\small NDC-scene~\cite{yao2023ndc}} & {\small OccFormer~\cite{zhang2023occformer}} \\
    \midrule
    \textbf{IoU}(\%) & \textbf{47.28} & \underline{46.21} & 44.58 & 44.15 & 37.24 & 36.50 \\
    \textbf{mIoU}(\%) & \textbf{17.14} & \underline{15.32} & 14.74 & 13.35 & 12.70 & 13.46 \\
    \midrule
    \textcolor[RGB]{91,155,213}{$\blacksquare$} \textbf{car}{\footnotesize (3.92\%)} & 
    \textbf{34.63} & \underline{33.31} & 27.33 & 26.54 & 26.26 & 25.09 \\
    \textcolor[RGB]{100,230,245}{$\blacksquare$} \textbf{bicycle}{\footnotesize (0.03\%)} & 
    0.53 & 0.61 & 1.07 & \underline{1.28} & \textbf{1.67} & 0.81 \\
    \textcolor[RGB]{30,60,150}{$\blacksquare$} \textbf{motorcycle}{\footnotesize (0.03\%)} & 
    \underline{1.69} & 0.46 & 1.14 & 0.56 & \textbf{2.37} & 1.19 \\
    \textcolor[RGB]{80,30,180}{$\blacksquare$} \textbf{truck}{\footnotesize (0.16\%)} & 
    \underline{20.12} & 6.03 & 17.06 & 7.26 & 14.75 & \textbf{25.53} \\
    \textcolor[RGB]{0,0,255}{$\blacksquare$} \textbf{other-veh.}{\footnotesize (0.20\%)} & 
    \textbf{11.06} & \underline{9.84} & 8.83 & 7.81 & 7.73 & 8.52 \\
    \textcolor[RGB]{255,30,30}{$\blacksquare$} \textbf{person}{\footnotesize (0.07\%)} & 
    0.54 & 0.47 & 2.25 & 1.93 & \textbf{3.60} & \underline{2.78} \\
    \textcolor[RGB]{255,37,199}{$\blacksquare$} \textbf{bicyclist}{\footnotesize (0.07\%)} & 
    0.16 & 0.10 & \textbf{4.09} & 1.97 & 2.74 & \underline{2.82} \\
    \textcolor[RGB]{150,30,90}{$\blacksquare$} \textbf{motorcyclist}{\footnotesize (0.05\%)} & 
    0.00 & 0.00 & 0.00 & 0.00 & 0.00 & 0.00 \\
    \textcolor[RGB]{255,0,255}{$\blacksquare$} \textbf{road}{\footnotesize (15.30\%)} & 
    \textbf{61.92} & 59.10 & 57.23 & 53.57 & \underline{59.20} & 58.85 \\
    \textcolor[RGB]{255,150,255}{$\blacksquare$} \textbf{parking}{\footnotesize (1.12\%)} & 
    \textbf{27.35} & 19.05 & 19.89 & 19.96 & \underline{21.42} & 19.61 \\
    \textcolor[RGB]{75,0,75}{$\blacksquare$} \textbf{sidewalk}{\footnotesize (11.13\%)} & 
    \textbf{33.97} & \underline{29.41} & 29.08 & 26.52 & 28.24 & 26.88 \\
    \textcolor[RGB]{175,0,75}{$\blacksquare$} \textbf{other-grnd.}{\footnotesize (0.56\%)} & 
    0.06 & 0.33 & \underline{1.26} & 0.42 & \textbf{1.67} & 0.31 \\
    \textcolor[RGB]{255,200,0}{$\blacksquare$} \textbf{building}{\footnotesize (14.10\%)} & 
    \textbf{25.89} & \underline{25.17} & 20.19 & 19.54 & 14.94 & 14.40 \\
    \textcolor[RGB]{255,120,50}{$\blacksquare$} \textbf{fence}{\footnotesize (3.90\%)} & 
    \textbf{10.13} & \underline{9.96} & 7.95 & 7.31 & 6.65 & 5.61 \\
    \textcolor[RGB]{0,175,0}{$\blacksquare$} \textbf{vegetation}{\footnotesize (39.3\%)} & 
    \textbf{30.08} & \underline{28.93} & 27.01 & 26.10 & 19.09 & 19.63 \\
    \textcolor[RGB]{135,60,0}{$\blacksquare$} \textbf{trunk}{\footnotesize (0.51\%)} & 
    \underline{9.54} & \textbf{9.58} & 7.71 & 6.10 & 3.51 & 3.93 \\
    \textcolor[RGB]{150,240,80}{$\blacksquare$} \textbf{terrain}{\footnotesize (9.17\%)} & 
    \textbf{38.51} & \underline{38.12} & 33.95 & 33.06 & 31.04 & 32.62 \\
    \textcolor[RGB]{255,240,150}{$\blacksquare$} \textbf{pole}{\footnotesize (0.29\%)} & 
    \underline{13.15} & \textbf{13.25} & 9.20 & 9.15 & 4.53 & 4.26 \\
    \textcolor[RGB]{255,0,0}{$\blacksquare$} \textbf{traf.-sign}{\footnotesize (0.08\%)} & 
    \underline{6.27} & \textbf{7.32} & 4.81 & 4.94 & 2.73 & 2.86 \\
    \bottomrule
  \end{tabular}
  }
  \label{tab:kitti-val}
\end{table*}

\subsubsection{Cubic Anisotropy}
Based on different adjacency relationships, the voxels within the $3\times 3\times 3$ cubic neighboring area can be categorized as follows:
\begin{itemize}
    \item 6 surface-adjacent voxels, denoted as $N_{\rm surface}$
    \item 12 edge-adjacent voxels, denoted as $N_{\rm edge}$
    \item 8 vertex-adjacent voxels, denoted as $N_{\rm vertex}$
\end{itemize}
For a specific voxel point $p$, the reassigned semantic differences against its neighboring voxels are calculated as follows:
\begin{align}
    \bm{S}_{\rm surface}^{p} &= \sum\limits_{n\in N_{\rm surface}^{p}} (v_{p} \oplus v_{n}) \; , \\
    \bm{S}_{\rm edge}^{p} &= \sum\limits_{n\in N_{\rm edge}^{p}} (v_{p} \oplus v_{n}) \; , \\
    \bm{S}_{\rm vertex}^{p} &= \sum\limits_{n\in N_{\rm vertex}^{p}} (v_{p} \oplus v_{n}) \; ,
\end{align}
where $v_{p}$ represents the reassigned voxel semantics of $p$ in $\bm{V}_{\rm re}$, and $\oplus$ is the exclusive or (XOR) operation:
\begin{equation}
    v_{p}\oplus v_{n} = \begin{cases}
    0, & v_{p}=v_{n} \; , \\ 
    1, & v_{p}\neq v_{n} \; .
    \end{cases}
\end{equation}
As demonstrated in Fig~\ref{sga}, considering only the surface-adjacency is insufficient to distinguish between voxels located along the object boundaries and those inside the object. Therefore, our Cubic Semantic Anisotropy (CSA) is calculated as a linear projection of the weighted sum of reassigned semantic differences, taking into account the different adjacency relationships within the cubic neighborhood:
\begin{equation}
    \bm{S}_{\rm CSA} = \alpha * \left(\bm{S}_{\rm surface}+w_{e}\cdot\bm{S}_{\rm edge}+w_{c}\cdot\bm{S}_{\rm vertex}\right) + \beta \; ,
\end{equation}
where $\alpha, \beta$ are the linear projection parameters, and $w_{e}, w_{v}$ are the weight parameters for edge- and vertex-adjacent reassigned semantic difference, respectively. The CSA enhances the cross entropy loss by introducing voxel-wise attention weights, allowing the model to focus more on the significant voxels:
\begin{equation}
    \mathcal{L}_{\rm CSA}(\bm{O}, \bm{V}) = -\dfrac{1}{N}\sum\limits_{n=1}^{N_v}\sum\limits_{c=1}^{N_c} S_{\rm CSA}^{n}\bm{O}_{n}\log \bm{V}_{n} \; ,
\end{equation}
where $N_v$ is the number of voxels for calculating the loss, $N_c$ is the number of target categories, and $\bm{O}, \bm{V}$ indicate the predicted occupancy and voxel labels, respectively.

It is important to emphasize that CSA only serves as auxiliary supervision guidance during training, and does not participate in the inference process, avoiding possible label leakage in evaluation.

\subsection{Critical Distribution Alignment}
An ideal 3D scene representation should maintain consistent feature distributions across different voxel grid resolutions. However, directly supervising the consistency over the entire scene is impractical due to the computational challenges posed by the vast number of voxels and the disturbance from large proportion of empty voxels. To address this issue, we introduce the Critical Distribution Alignment (CDA) module, which effectively promotes the feature distribution consistency by focusing on the selected critical voxels.

\subsubsection{Critical Voxel Selection}
HASSC~\cite{wang2024not} proposes a global hard voxel mining strategy based on prediction errors, which selects confusing voxels that are prone to misclassifications, failing to represent the semantic distribution throughout the scene. In contrast, we identify critical voxels at each resolution level with the guidance of occupancy confidence and cubic semantic anisotropy, focusing on contextual semantic significance. 
Specifically, the 3D features with resolution level $i$ are fed into an occupancy head with MLP layers to generate the occupancy prediction results:
\begin{equation}
    \bm{O}_{i} = {\rm MLP}(\bm{F}_{i}^{\rm 3D}) \; ,
\end{equation}
where $\bm{O}_{i}\in\mathbb{R}^{X_i\times Y_i\times Z_i\times N}$ represents the probability distribution of each voxels across the $N$ target semantic classes. For each voxel, we adopt the highest predicted probability among the $N$ classes as its occupancy confidence:
\begin{equation}
    \bm{C}_{i}^{\rm occ} = {\rm max}\left(\bm{O}_{i}, {\rm dim}=-1\right) \; .
\end{equation}
The critical voxels are selected as the top $k$ voxels with the largest product of occupancy confidence and the cubic semantic anisotropy:
\begin{equation}
    \bm{V}_{i}^{\rm critic} = {\rm top}_{k}\left(\bm{C}_{i}^{\rm occ}\cdot \bm{S}_{\rm CSA}\right) \; ,
\end{equation}
where $\bm{V}_{i}^{\rm critic}\in\mathbb{R}^{k\times C}$ denotes the selected $k$ critical voxel features, and $k$ is the voxel number threshold.

\subsubsection{Circulated Loss}
Guided by occupancy confidence and cubic semantic anisotropy, these critical voxels carry both discriminative semantics and semantic-aware significance.
Therefore, we employ the critical voxels as instance-level anchors with representative feature distributions over the entire scene.
Notice that voxel at $(x, y, z)$ in the low resolution grid softly corresponds to the voxel at $(\lambda x, \lambda y, \lambda z)$ in the high resolution grid, where $\lambda = X_1 / X_2$ is the ratio between high and low resolutions, the critical voxels across different resolutions should possess consistent feature distributions.
To supervise the consistency between the critical feature distributions at different resolutions $\bm{V}_{1}^{\rm critical}, \bm{V}_{2}^{\rm critical}$, we propose the following Circulated Loss:
\begin{equation}
    \mathcal{L}_{\rm circ} = D_{\rm KL}(\bm{V}_{1}^{\rm critic} || \bm{V}_{2}^{\rm critic}) + D_{\rm KL}(\bm{V}_{2}^{\rm critic} || \bm{V}_{1}^{\rm critic}) \; ,
\end{equation}
where $D_{\rm KL}$ is the Kullback-Leibler divergence.
The above alignment acts as a soft constraint to encourage contextual consistency across different resolutions, without forcing strict voxel-wise one-to-one correspondence.

\begin{table*}
  \centering
  \caption{Semantic scene completion comparison results against the state-of-the-art camera-based methods on the SemanticKITTI~\cite{behley2019semantickitti} hidden test set. Best results are highlighted in \textbf{bold}, and second-best results are \underline{underlined}.}
  \resizebox{1.0\linewidth}{!}{
  \begin{tabular}{@{}l|cccccc@{}}
    \toprule
    \textbf{Method} & {\small MRA~(Ours)} & {\small SGN~\cite{mei2024camera}} & {\small Bi-SSC~\cite{xue2024bi}} & {\small NDC-scene~\cite{yao2023ndc}} & {\small OccFormer~\cite{zhang2023occformer}} & {\small TPVFormer~\cite{huang2023tri}} \\
    \midrule
    \textbf{IoU}(\%) & \textbf{46.65} & \underline{45.42} & 45.10 & 36.19 & 34.53 & 34.25 \\
    \textbf{mIoU}(\%) & \textbf{16.82} & 15.76 & \underline{16.73} & 12.58 & 12.32 & 11.26 \\
    \midrule
    \textcolor[RGB]{91,155,213}{$\blacksquare$} \textbf{car}{\footnotesize (3.92\%)} & 
    \textbf{26.90} & \underline{25.40} & 25.00 & 19.13 & 21.60 & 19.20 \\
    \textcolor[RGB]{100,230,245}{$\blacksquare$} \textbf{bicycle}{\footnotesize (0.03\%)} & 
    0.60 & 0.90 & \underline{1.80} & \textbf{1.93} & 1.50 & 1.00 \\
    \textcolor[RGB]{30,60,150}{$\blacksquare$} \textbf{motorcycle}{\footnotesize (0.03\%)} & 
    1.00 & 1.60 & \textbf{2.90} & \underline{2.07} & 1.70 & 0.50 \\
    \textcolor[RGB]{80,30,180}{$\blacksquare$} \textbf{truck}{\footnotesize (0.16\%)} & 
    3.10 & 4.50 & \textbf{6.80} & \underline{4.77} & 1.20 & 3.70 \\
    \textcolor[RGB]{0,0,255}{$\blacksquare$} \textbf{other-veh.}{\footnotesize (0.20\%)} & 
    6.00 & 3.70 & \textbf{6.80} & \underline{6.69} & 3.20 & 2.30 \\
    \textcolor[RGB]{255,30,30}{$\blacksquare$} \textbf{person}{\footnotesize (0.07\%)} & 
    1.30 & 0.50 & 1.70 & \textbf{3.44} & \underline{2.20} & 1.10 \\
    \textcolor[RGB]{255,37,199}{$\blacksquare$} \textbf{bicyclist}{\footnotesize (0.07\%)} & 
    2.00 & 0.30 & \textbf{3.30} & \underline{2.77} & 1.10 & 2.40 \\
    \textcolor[RGB]{150,30,90}{$\blacksquare$} \textbf{motorcyclist}{\footnotesize (0.05\%)} & 
    0.00 & 0.10 & \underline{1.00} & \textbf{1.64} & 0.20 & 0.30 \\
    \textcolor[RGB]{255,0,255}{$\blacksquare$} \textbf{road}{\footnotesize (15.30\%)} & 
    \underline{61.2} & 60.40 & \textbf{63.40} & 58.12 & 55.90 & 55.10 \\
    \textcolor[RGB]{255,150,255}{$\blacksquare$} \textbf{parking}{\footnotesize (1.12\%)} & 
    \textbf{33.20} & 28.90 & \underline{31.70} & 25.31 & 31.50 & 27.40 \\
    \textcolor[RGB]{75,0,75}{$\blacksquare$} \textbf{sidewalk}{\footnotesize (11.13\%)} & 
    \textbf{33.70} & 31.40 & \underline{33.30} & 28.05 & 30.30 & 27.20 \\
    \textcolor[RGB]{175,0,75}{$\blacksquare$} \textbf{other-grnd.}{\footnotesize (0.56\%)} & 
    \underline{9.90} & 8.70 & \textbf{11.20} & 6.53 & 6.50 & 6.50 \\
    \textcolor[RGB]{255,200,0}{$\blacksquare$} \textbf{building}{\footnotesize (14.10\%)} & 
    \textbf{29.30} & \underline{28.40} & 26.60 & 14.90 & 15.70 & 14.80 \\
    \textcolor[RGB]{255,120,50}{$\blacksquare$} \textbf{fence}{\footnotesize (3.90\%)} & 
    \underline{19.20} & 18.10 & \textbf{19.40} & 12.85 & 11.90 & 11.00 \\
    \textcolor[RGB]{0,175,0}{$\blacksquare$} \textbf{vegetation}{\footnotesize (39.3\%)} & 
    \textbf{29.50} & \underline{27.40} & 26.10 & 17.94 & 16.80 & 13.90 \\
    \textcolor[RGB]{135,60,0}{$\blacksquare$} \textbf{trunk}{\footnotesize (0.51\%)} & 
    \textbf{13.70} & \underline{12.60} & 10.50 & 3.49 & 3.90 & 2.60 \\
    \textcolor[RGB]{150,240,80}{$\blacksquare$} \textbf{terrain}{\footnotesize (9.17\%)} & 
    \textbf{30.40} & 28.40 & \underline{28.9} & 25.01 & 21.30 & 20.40 \\
    \textcolor[RGB]{255,240,150}{$\blacksquare$} \textbf{pole}{\footnotesize (0.29\%)} & 
    \textbf{10.30} & \underline{10.00} & 9.30 & 4.43 & 3.80 & 2.90 \\
    \textcolor[RGB]{255,0,0}{$\blacksquare$} \textbf{traf.-sign}{\footnotesize (0.08\%)} & 
    8.20 & \underline{8.30} & \textbf{8.40} & 2.96 & 3.70 & 1.50 \\
    \bottomrule
  \end{tabular}
  }
  \label{tab:kitti-test}
\end{table*}

\subsection{Training and Inference}
\subsubsection{Training Loss} Following~\cite{cao2022monoscene, mei2024camera}, we adopt the CSA-enhanced cross entropy loss~(where CSA serves as training weights for auxiliary supervision guidance), lovasz loss~\cite{berman2018lovasz}, and scene-class affinity loss to supervise the multi-resolution occupancy prediction results:
\begin{equation}
\begin{aligned}
    \mathcal{L}_{\rm occ}^{i} &= \mathcal{L}_{\rm CSA}(\bm{O}_i, \bm{V}_{\rm gt}^{i}) + \mathcal{L}_{\rm lovasz}(\bm{O}_i, \bm{V}_{\rm gt}^{i}) \\ 
    &+ \mathcal{L}_{scal}(\bm{O}_i, \bm{V}_{\rm gt}^{i}) \; ,
\end{aligned}
\end{equation}
where $\bm{V}_{\rm gt}^{i}$ is the ground truth voxel semantic labels rescaled to resolution level $i$. The overall objective function is formulated as follows:
\begin{equation}
    \mathcal{L_{\rm ssc}} = L_{\rm occ}^{1} + L_{\rm occ}^{2} + \gamma\cdot L_{\rm circ} \; ,
\end{equation}
where the Circulated Loss $L_{\rm circ}$ serves as complementary supervision signals to the occupancy loss, addressing voxel sparsity by promoting self-alignment across multi-resolution 3D features. Furthermore, we also employ the self-distillation~\cite{he2020momentum, wang2024not} training strategy as the model-level self-alignment for improved accuracy and robustness.

\subsubsection{Inference} 
The student-branch is well optimized during training to address the voxel sparsity through self-alignment across multi-resolution 3D features, while also leveraging the soft label provided by the teacher branch. During inference, only the student-branch is retained and the high-resolution occupancy prediction is utilized as the final output, eliminating additional computational overhead. Since no loss computation is needed during inference, the CSA module is not activated, avoiding label leakage in evaluation.

\section{Experiments}
\label{sec:experiment}

In this section, we describe the dataset, evaluation metrics, and the implementation details of our approach. Subsequently, we conduct extensive experiments to demonstrate that our MRA approach consistently outperforms the state-of-the-art methods on the complex large-scale outdoor dataset SemanticKITTI~\cite{behley2019semantickitti} as well as SSCBench-KITTI-360~\cite{li2023sscbench}. Furthermore, we conduct comprehensive ablation experiments and qualitative results, thereby offering an in-depth analysis on the efficacy of our MRA approach and effectiveness of different architectural components.

\begin{table*}
  \centering
  \caption{Semantic scene completion comparison results against the state-of-the-art camera-based methods on the SSCBench-KITTI360~\cite{behley2019semantickitti} test set. Best results are highlighted in \textbf{bold}, and second-best results are \underline{underlined}.}
  \resizebox{1.0\linewidth}{!}{
  \begin{tabular}{@{}l|cccccc@{}}
    \toprule
    \textbf{Method} & {\small MRA~(Ours)} & {\small SGN~\cite{mei2024camera}} & {\small Symphonies~\cite{jiang2024symphonize}} & {\small DepthSSC~\cite{yao2023depthssc}} & {\small OccFormer~\cite{zhang2023occformer}} & {\small VoxFormer~\cite{li2023voxformer}} \\
    \midrule
    \textbf{IoU}(\%) & \textbf{48.03} & \underline{47.06} & 44.12 & 40.85 & 40.27 & 38.76 \\
    \textbf{mIoU}(\%) & \textbf{20.14} & 18.25 & \underline{18.58} & 14.28 & 13.81 & 11.91 \\
    \midrule
    \textcolor[RGB]{91,155,213}{$\blacksquare$} \textbf{car}{\footnotesize (2.85\%)} & 
    \textbf{33.97} & 29.03 & \underline{30.02} & 21.90 & 22.58 & 17.84 \\
    \textcolor[RGB]{100,230,245}{$\blacksquare$} \textbf{bicycle}{\footnotesize (0.01\%)} & 
    0.00 & \textbf{3.43} & 1.85 & \underline{2.36} & 0.66 & 1.16 \\
    \textcolor[RGB]{30,60,150}{$\blacksquare$} \textbf{motorcycle}{\footnotesize (0.01\%)} & 
    2.67 & 2.90 & \textbf{5.90} & \underline{4.30} & 0.26 & 0.89 \\
    \textcolor[RGB]{80,30,180}{$\blacksquare$} \textbf{truck}{\footnotesize (0.16\%)} & 
    \textbf{25.89} & 10.89 & \underline{25.07} & 11.51 & 9.89 & 4.56 \\
    \textcolor[RGB]{0,0,255}{$\blacksquare$} \textbf{other-veh.}{\footnotesize (5.75\%)} & 
    \textbf{13.11} & 5.20 & \underline{12.06} & 4.56 & 3.80 & 2.06 \\
    \textcolor[RGB]{255,30,30}{$\blacksquare$} \textbf{person}{\footnotesize (0.02\%)} & 
    \underline{4.01} & 2.99 & \textbf{8.20} & 2.92 & 2.77 & 1.63 \\
    \textcolor[RGB]{255,0,255}{$\blacksquare$} \textbf{road}{\footnotesize (14.98\%)} & 
    \textbf{61.19} & \underline{58.14} & 54.94 & 50.88 & 54.30 & 47.01 \\
    \textcolor[RGB]{255,150,255}{$\blacksquare$} \textbf{parking}{\footnotesize (2.31\%)} & 
    13.21 & \textbf{15.04} & \underline{13.83} & 12.89 & 13.44 & 9.67 \\
    \textcolor[RGB]{75,0,75}{$\blacksquare$} \textbf{sidewalk}{\footnotesize (6.43\%)} & 
    \textbf{36.82} & \underline{36.40} & 32.76 & 30.27 & 31.53 & 27.21 \\
    \textcolor[RGB]{175,0,75}{$\blacksquare$} \textbf{other-grnd.}{\footnotesize (2.05\%)} & 
    \textbf{8.85} & 4.43 & \underline{6.93} & 2.49 & 3.55 & 2.89 \\
    \textcolor[RGB]{255,200,0}{$\blacksquare$} \textbf{building}{\footnotesize (15.67\%)} & 
    \underline{37.55} & \textbf{42.02} & 35.11 & 37.33 & 36.42 & 31.18 \\
    \textcolor[RGB]{255,120,50}{$\blacksquare$} \textbf{fence}{\footnotesize (0.96\%)} & 
    \textbf{9.13} & 7.72 & \underline{8.58} & 5.22 & 4.80 & 4.97 \\
    \textcolor[RGB]{0,175,0}{$\blacksquare$} \textbf{vegetation}{\footnotesize (41.99\%)} & 
    \textbf{41.37} & 38.17 & \underline{38.33} & 29.61 & 31.00 & 28.99 \\
    \textcolor[RGB]{150,240,80}{$\blacksquare$} \textbf{terrain}{\footnotesize (7.10\%)} & 
    13.99 & \textbf{23.22} & 11.52 & \underline{21.59} & 19.51 & 14.69 \\
    \textcolor[RGB]{255,240,150}{$\blacksquare$} \textbf{pole}{\footnotesize (0.22\%)} & 
    \textbf{17.87} & \underline{16.73} & 14.01 & 5.97 & 7.77 & 6.51 \\
    \textcolor[RGB]{255,0,0}{$\blacksquare$} \textbf{traf.-sign}{\footnotesize (0.06\%)} & 
    \underline{14.92} & \textbf{16.38} & 9.57 & 7.71 & 8.51 & 6.92 \\
    \textcolor[RGB]{250,150,0}{$\blacksquare$} \textbf{other-struct.}{\footnotesize (4.33\%)} & 
    \textbf{15.63} & 9.93 & \underline{14.44} & 5.24 & 6.95 & 3.79 \\
    \textcolor[RGB]{50,255,255}{$\blacksquare$} \textbf{other-obj.}{\footnotesize (0.28\%)} & 
    \textbf{12.29} & 5.86 & \underline{11.28} & 3.51 & 4.60 & 2.43 \\
    \bottomrule
  \end{tabular}
  }
  \label{tab:kitti360-test}
\end{table*}

\subsection{Dataset and Metrics}
\subsubsection{Dataset} For large-scale outdoor scene understanding, the KITTI odometry dataset~\cite{geiger2012we} comprises 22 sequences across 20 classes cpatuered using a Velodyne HDL-64 laser scanner in autonomous driving environments. The \textbf{SemanticKITTI}~\cite{behley2019semantickitti} dataset, built upon the KITTI dataset, provides semantic annotations for all sequences. According to the official setting for the 3D semantic  scene completion (SSC) task, sequences 00-07 and 09-10 (a total of 3834 scans) are designated for training, sequence 08 (815 scans) is used for validation, and the remaining sequences (3901 scans) are for testing. The \textbf{SSCBench-KITTI-360}~\cite{li2023sscbench} benchmark features nine densely annotated sequences of urban driving scenes for semantic scene completion. The training set includes 8,487 frames from scenes 00, 02-05, 07 and 10, while the validation set consists of 1,812 frames from scene 06. The testing set contains 2,566 frames exclusively from scene 09, providing a robust and diverse environment for model evaluation. Both SSC benchmarks focus on the 3D volumes within the range of $[0 \sim 51.2m, -25.6m \sim 25.6m, -2 \sim 4.4m]$, with a voxel resolution of $0.2m$. The SSC labels for training and validation are with the resolution of $256\times 256\times 32$.

\subsubsection{Metrics} Following~\cite{li2023voxformer, mei2024camera}, we employ the Intersection over Union (IoU) of occupied voxels as the evaluation metric for the task of class-agnostic scene scene (SC). Additionally, we report the mean Intersection over Union (mIoU) across all semantic categories to measure the performance of the semantic scene completion (SSC):
\begin{equation}
\begin{aligned}
    {\rm IoU} &= \dfrac{TP}{TP+FP+FN} \; , \\
    {\rm mIoU} &= \dfrac{1}{C}\sum\limits_{c=1}^{C}\dfrac{TP_c}{TP_{c}+FP_{c}+FN_{c}} \; ,
\end{aligned}
\end{equation}
where $TP, FP, FN$ represent the number of true positive, false positive, and false negative occupancy predictions, and $C$ stands for the total number of classes.

\begin{table}[t]
    \centering
    \caption{Quantitative comparison results in different ranges on the SemanticKITTI validation set. Best results are highlighted in \textbf{bold}, and second-best results are \underline{underlined}.}
    \resizebox{1.0\columnwidth}{!}{
    \begin{tabular}{l|ccc|ccc}
    \toprule
    \multirow{2}{*}{\textbf{Methods}} & \multicolumn{3}{c|}{\textbf{IoU}(\%)} & \multicolumn{3}{c}{\textbf{mIoU}(\%)} \\
     & 12.8m & 25.6m & 51.2m &  12.8m & 25.6m & 51.2m \\
    \midrule
    MonoScene~\cite{cao2022monoscene} & 38.42 & 38.55 & 36.80 & 12.25 & 12.22 & 11.30 \\
    OccFormer~\cite{zhang2023occformer} & 56.38 & 47.28 & 36.50 & 20.91 & 17.90 & 13.46 \\
    VoxFormer~\cite{li2023voxformer} & 65.38 & 57.69 & 44.15 & 21.55 & 18.42 & 13.35 \\
    HASSC~\cite{wang2024not} & 66.05 & 58.01 & 44.58 & 24.10 & 20.27 & 14.74 \\
    SGN~\cite{mei2024camera} & \underline{70.61} & \underline{61.90} & \underline{46.21} & \underline{25.70} & \underline{22.02} & \underline{15.32} \\
    \midrule
    MRA~(Ours) & \textbf{71.36} & \textbf{62.69} & \textbf{47.28} & \textbf{27.67} & \textbf{24.10} & \textbf{17.14} \\
    \bottomrule
    \end{tabular}
    }
    \label{tab:ranges}
\end{table}

\subsection{Implementation Details}
We crop the input RGB images from cam2 to the size of $1220\times 370$ for the SemanticKITTI dataset and input RGB images from cam1 to the size of $1408\times 376$ for the SSCBench-KITTI-360 dataset. The length of temporal inputs is set as 5. The 2D feature maps, downsampled to 1/16 of the original input resolution, are utilized for subsequent processing and thhe feature dimension $C$ is set to 128. In MVT, we set two resolution levels where the high resolution $X_1\times Y_1\times Z_1$ of 3D feature volume is $128\times 128\times 16$ and the low resolution $X_2\times Y_2\times Z_2$ is $64\times 64\times 8$. The final predictions are upsampled to $256\times 256\times 32$. In CSA, the weight parameters of semantic difference aggregation and linear projection are set as $w_e=0.1, w_v=0.3, \alpha=1.0, \beta=0.5$ respectively. In CDA, we select $k=4096$ critical voxels and the weight of the Circulated Loss is $\gamma=1.0$. We train our MRA model for 24 epochs on 4 A6000 GPUs with a total batch size of 4. The AdamW\cite{loshchilov2017decoupled} optimizer is used with an initial learning rate of 2e-4 and a weight decay of 1e-2.

\begin{figure*}[ht]
\centering
\includegraphics[width=1.0\textwidth]{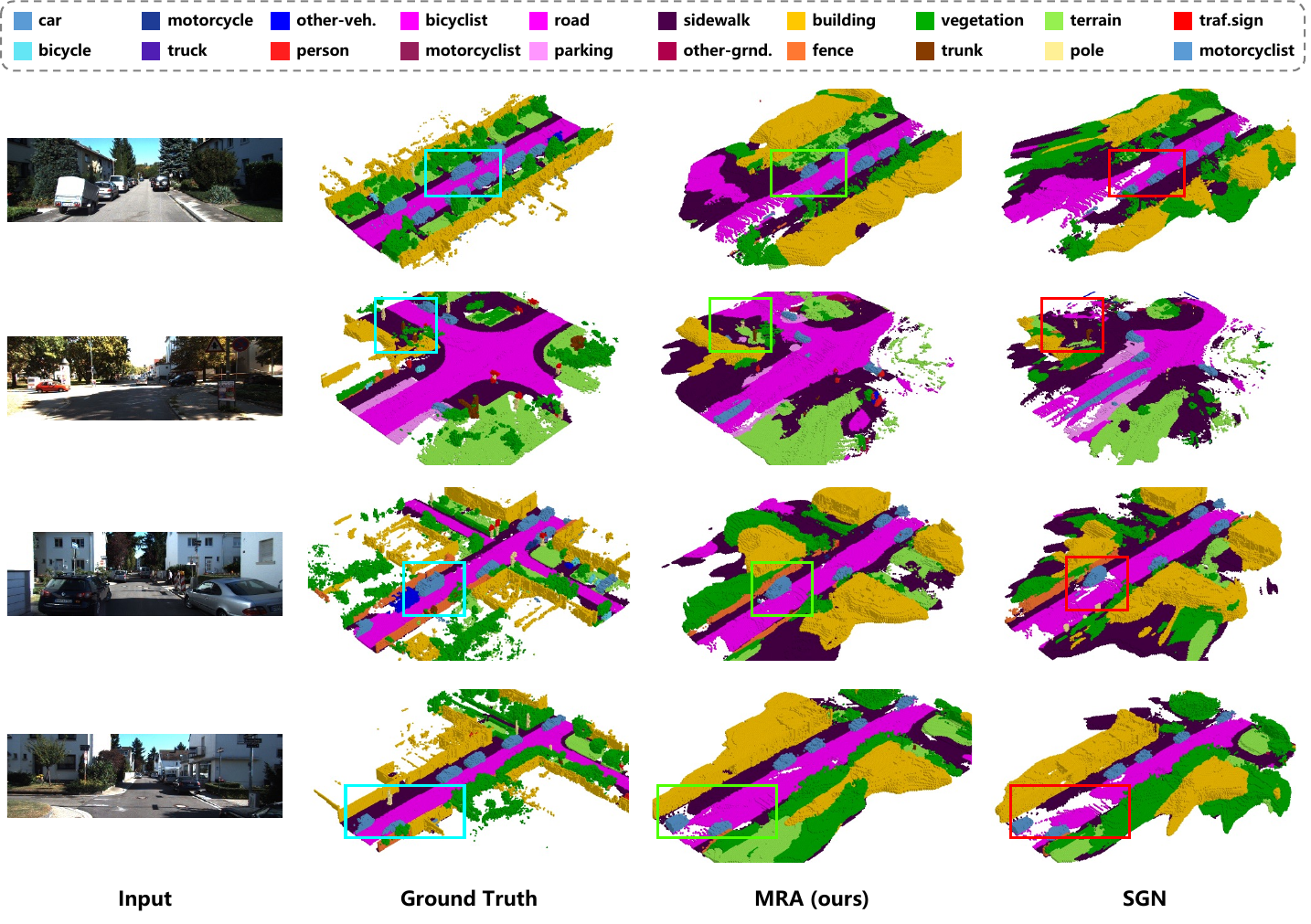} 
\caption{Qualitative visualization results on the SemanticKITTI validation set. Cyan boxes outline the occupancy ground truth. Red boxes indicate false occupancy predictions of the best comparison method SGN, and green boxes indicate the improved scene completion results with more accurate object boundaries generated by our MRA approach. Better viewed when zoomed in.}
\label{visual}
\end{figure*}

\subsection{Comparison With the State-of-the-Art}
We conduct comparison experiments on two popular SemanticKITTI and SSCBench-KITTI-360 benchmarks to validate the effectiveness of our proposed MRA method.
\begin{itemize}
    \item \textbf{SemanticKITTI}: Table~\ref{tab:kitti-val} demonstrates the comparison results between our MRA approach and other state-of-the-art methods on the SemanticKITTI validation dataset, where the best results are highlighted in bold and the second-best results are underlined. It can be observed that our MRA approach achieves superior performance of \textbf{47.28\%} IoU for class-agnostic scene completion (SC) and \textbf{17.14\%} mIoU for semantic scene completion (SSC), with performance improvements of \textbf{1.07\%} IoU and \textbf{1.82\%} mIoU over the best comparison methods SGN~\cite{mei2024camera}, respectively. We also present the comparison results on the more challenging SemanticKITTI test dataset in Table~\ref{tab:kitti-test}, where our MRA approach still achieve the best performance of \textbf{46.65\%} IoU and \textbf{16.82\%} mIoU, respectively.
\begin{figure*}[ht]
\centering
\includegraphics[width=1.0\textwidth]{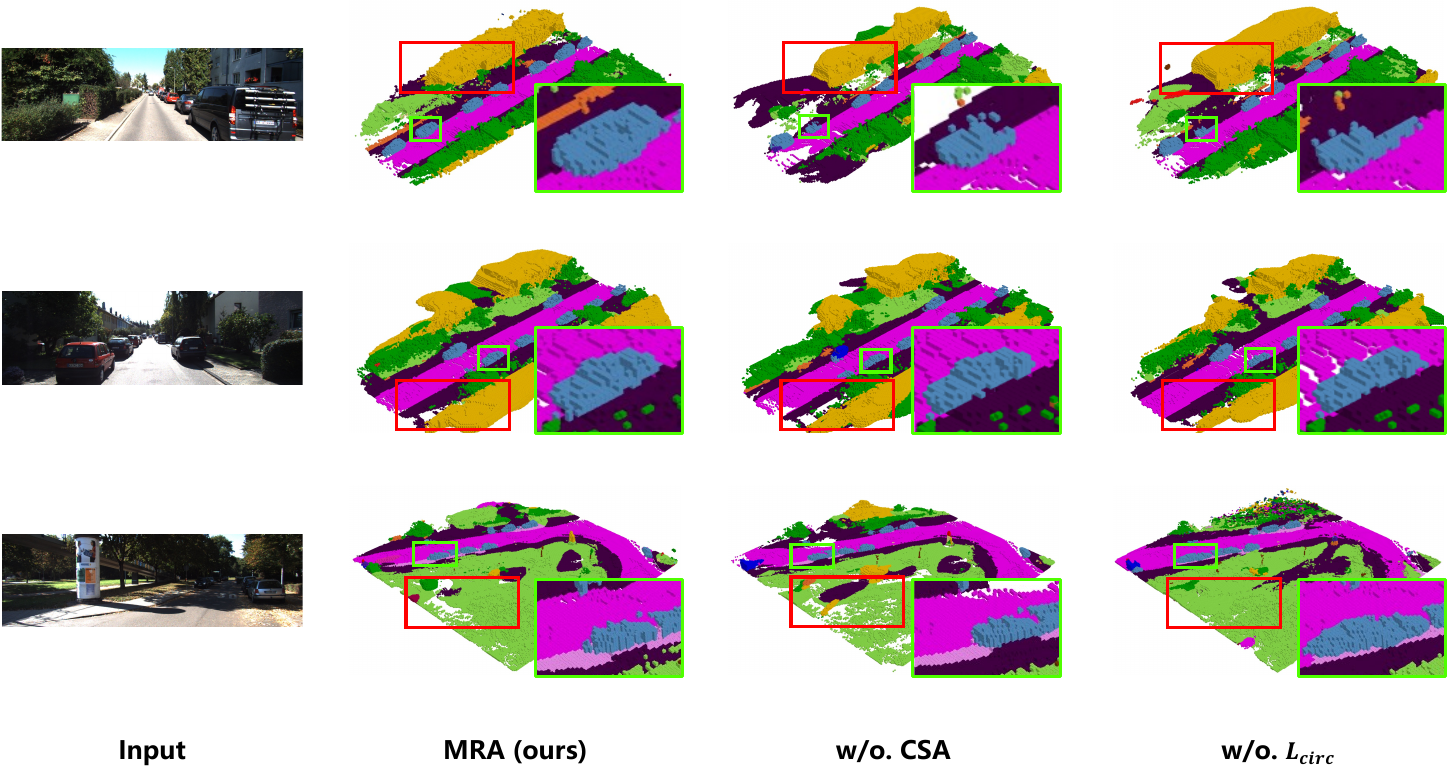} 
\caption{Qualitative visualization results among MRA, w/o. CSA and w/o. $L_{\rm circ}$ settings. Green boxes highlight examples where MRA achieves sharper and more accurate object boundaries compared to the ablated versions. Red boxes indicate failure cases where MRA struggles, particularly in distant areas with small-scale objects and occlusion issues. Better viewed when zoomed in.}
\label{visual_ablation}
\end{figure*}
    \item \textbf{SSCBench-KITTI-360}: Table~\ref{tab:kitti360-test} presents the comparison results of our MRA approach against other state-of-the-art methods on the SSCBench-KITTI-360 test dataset. The best results are highlighted in bold, while the second-best results are underlined. It can be observed that our MRA approach achieves superior performance of \textbf{48.03\%} IoU for the SC task and \textbf{20.14\%} mIoU for the SSC task, with performance improvements of \textbf{0.97\%} IoU and \textbf{1.89\%} mIoU over the best comparison methods SGN~\cite{mei2024camera} and Symphonies~\cite{jiang2024symphonize}, respectively. The consistent and significant performance gains on both datasets highlight the effectiveness and generalization capability of our MRA approach.
    \item Compared with existing methods, the performance improvements of our MRA approach are attributed to the auxiliary supervision signals provided by the self-alignment across multi-resolution 3D features. SGN~\cite{mei2024camera} employs the dense-sparse-dense framework with hybrid geometric and semantic guidance to exploit discriminative 3D features and propagate fine-grained semantics for 3D semantic scene completion. However, the inherent voxel sparsity of autonomous driving scenarios harms the quality of generated geometric and semantic guidance, thereby limiting the extraction of discriminative features and the convergence of semantic propagation to the whole scene. On the contrary, our MRA approach not only locates seed voxels with depth-aware global proposals, but further improves discriminative 3D features with the fusion and alignment across different resolution levels as well. As a result, the aligned multi-resolution 3D features facilitate more accurate scene understanding with consistent semantics throughout the whole scene, leading to improved 3D semantic scene completion performance. Symphonies~\cite{jiang2024symphonize} proposes the serial instance-propagated attentions to integrate instance-level information for 2D-to-3D view transformation and 3D scene completion, but the optimization of instance queries is still solely depended on the sparse voxel labels, limiting the overall accuracy of the model performance. By contrast, our MRA approach selects critical levels with respect to the occupancy confidence and cubic semantic anisotropy. Therefore, the selected critical voxels with both discriminative features and semantic-aware significance serve as instance-level anchors of the entire 3D scene, and promotes the scene representation through the critical distribution alignment supervised by the circulate loss.
    \item \textbf{Comparison in Different Ranges}: Table~\ref{tab:ranges} illustrates the comparison results in different ranges of 12.8m, 25.6m, 51.2m, respectively. It can be observed that our MRA approach surpasses the comparison methods in all ranges, with performance improvements of 0.75\% IoU and 1.97\% mIoU within 12.8 meters over the second best results, as well as 0.79\% IoU and 2.08\% mIoU performance gains within 25.6 meters respectively. The results showcase the effectiveness and robustness of our MRA approach.
\end{itemize}

\subsection{Qualitative results.} 
Figure \ref{visual} presents the qualitative visualization results of the proposed MRA approach on the SemanticKITTI validation set. The first row demonstrates the input images. The second to fourth rows present the visualization results of the ground truth, our MRA approach, and the best comparison method SGN~\cite{mei2024camera}, respectively. We highlight the occupancy ground truth with cyan boxes in the second row as a reference for comparison. The red boxes in the fourth row indicate the area of false voxel predictions from SGN, while the green boxes in the third row highlight the improved scene completion results from our MRA approach, completing missing objects such as ``car" and specific areas of ``road", as well as correcting the false predictions across similar categories such as ``road" and ``sidewalk". 
This qualitative analysis underscores the efficacy of our MRA approach in addressing voxel sparsity with self-alignment between multi-resolution 3D features, generating more accurate scene completion results.
Figure~\ref{visual_ablation} further presents qualitative visualization results among MRA, w/o. CSA, and w/o. $L_{\rm circ}$ settings. Green boxes highlight examples where MRA achieves sharper and more accurate object boundaries, demonstrating the effectiveness of CSA and $L_{\rm circ}$ in improving object boundary predictions with contextual semantic significance identification and global critical distributional alignment. Red boxes indicate failure cases where MRA struggles, particularly in distant areas with small-scale objects and occlusion issues, where only low-quality semantic features are available, leading to incomplete and false semantic scene completion results.

\begin{table}[t]
    \centering
    \caption{Ablation study on the SemanticKITTI validation set validating the effectiveness of  different architectural components of our MRA approach.}
    \begin{tabular}{ccccc|c|c}
    \toprule
    Variants & \textbf{MVT} & \textbf{CSA} & \textbf{CDA} & \textbf{SD} & \textbf{IoU}(\%) & \textbf{mIoU}(\%) \\
    \midrule
     baseline & & & & & 44.15 & 13.35 \\
     1 & $\checkmark$ & & & & 44.74 & 14.28 \\
     2 & $\checkmark$ & $\checkmark$ & & & 45.53 & 15.78 \\
     3 & $\checkmark$ & & $\checkmark$ & & 45.68 & 15.87 \\
     4 & $\checkmark$ & $\checkmark$ & $\checkmark$ & & 46.93 & 16.74 \\
     5 & $\checkmark$ & $\checkmark$ & $\checkmark$ & $\checkmark$ & \textbf{47.28} & \textbf{17.14} \\
    \bottomrule
    \end{tabular}
    \label{tab:ablation}
\end{table}
\vspace{-5mm}

\subsection{Ablation Studies}

\subsubsection{Ablation on Network Components}
As illustrated in Fig~\ref{motivation}, our MRA approach consists of the multi-resolution view transformer (MVT) module, cubic semantic anisotropy (CSA) module, critical distribution alignment (CDA) module, and self-distillation (SD) training strategy. In this section, we further conduct ablation experiments on the SemanticKITTI validation dataset to investigate the effectiveness of the designed modules, as demonstrated in Table~\ref{tab:ablation}. It can be observed that:

\begin{table}[t]
    \centering
    \caption{Ablation study on the SemanticKITTI validation set for the weight parapets of semantic differences aggregation in cubic semantic anisotropy.}
    \begin{tabular}{cc|c|c}
    \toprule
    \textbf{Edge Weight} $w_e$ & \textbf{Vertex Weight} $w_v$ & \textbf{IoU}(\%) & \textbf{mIoU}(\%) \\
    \midrule
    0.1 & 0.1 & \textbf{45.58} & 15.34 \\
    0.1 & 0.3 & 45.53 & \textbf{15.78} \\
    0.1 & 0.5 & \textbf{45.58} & 15.56 \\
    0.3 & 0.1 & 45.51 & 15.26 \\
    0.3 & 0.3 & 45.50 & 15.51 \\
    0.3 & 0.5 & 45.45 & 15.48 \\
    \bottomrule
    \end{tabular}
    \label{tab:ablation-csa-weight}
\end{table}

\begin{itemize}
    \item We adopt VoxFormer~\cite{li2023voxformer} as the baseline method, which achieves $44.15\%$ IoU and $13.5\%$ mIoU performance for the scene completion (SC) and semantic scene completion (SSC) task respectively. By adding the multi-resolution view transformer (MVT) module to the baseline, the model performances are improved by $0.59\%$ IoU and $0.93\%$ mIoU respectively, which showcases the effectiveness of globally aligning multi-resolution 3D features with the fusion of seed features and the propagation of discriminative semantics.
    \item Based on MVT, we further equip the model with the cubic semantic anisotropy (CSA) module and ritical distribution alignment (CDA) module, respectively. The CSA module brings $0.79\%$ IoU and $1.50\%$ mIoU performance improvements via identifying semantic-aware significance of each voxel so that voxels with discriminative semantics gain more attention, thereby alleviating the disturbance of large portion empty voxels. The CDA modules improves the semantic scene completion performance by $0.94\%$ IoU and $1.59\%$ mIoU with the local instance-level alignment across critical distributions of multi-resolution 3D features, facilitating scene comprehension with more accurate details.
    \item By incorporating MVT, CSA, and CDA together, our MRA approach execute both global and local alignment across multi-resolution 3D features, providing auxiliary supervision signals for addressing the challenge of voxel sparsity. The synergistic effect of MVT, CSA, and CDA leads to significant performance improvements of $2.78\%$ IoU and $3.29\%$ mIoU over the baseline methods. Furthermore, when adopting the self-distillation training strategy, we obtain the best performance of $47.28\%$ IoU for scene completion and $17.14\%$ mIoU for semantic scene completion, respectively.
    \item To further investigate different choices of feature projection mechanisms, Table~\ref{tab:ablation-flosp} presents the comparison results employing cross attention~(CA) and FlosP with ASPP for feature projection, respectively. It can be observed that FLosP consistently achieves slightly better performance under baseline, MVT, and MRA configurations, which aligns with our efficient and flexible architectural design rationale. Therefore, we utilize FLosP with ASPP for feature projection in our MRA framework.
\end{itemize}

\begin{table}[t]
    \centering
    \caption{Ablation study on the SemanticKITTI validation set for different feature projection mechanisms.}
    \begin{tabular}{c|cc|cc|cc}
    \toprule
    \multirow{2}{*}{Variants} & \multicolumn{2}{c|}{baseline} & \multicolumn{2}{c|}{MVT} & \multicolumn{2}{c}{MRA} \\
    & CA & FLosP & CA & FLosP & CA & FLosP \\
    \midrule
    \textbf{IoU}(\%) & 44.15 & 44.26 & 44.71 & 44.74 & 47.16 & 47.28 \\
    \textbf{mIoU}(\%) & 13.35 & 13.49 & 14.26 & 14.28 & 17.12 & 17.14 \\
    \bottomrule
    \end{tabular}
    \label{tab:ablation-flosp}
\end{table}

\begin{table}[t]
    \centering
    \caption{Ablation study on the SemanticKITTI validation set for linear projection parameters in cubic semantic anisotropy.}
    \begin{tabular}{c|ccccc}
    \toprule
    $\beta (\alpha=1)$ & 0.1 & 0.3 & 0.5 & 0.7 & 0.9 \\
    \midrule
    \textbf{IoU}(\%) & 45.19 & 45.35 & \textbf{45.53} & 45.26 & 45.44 \\
    \textbf{mIoU}(\%) & 15.46 & 15.59 & \textbf{15.78} & 15.64 & 15.30 \\
    \bottomrule
    \end{tabular}
    \label{tab:ablation-csa-proj}
\end{table}

\begin{table}[t]
    \centering
    \caption{Ablation study on the SemanticKITTI validation set for different configurations of the semantic groups in cubic semantic anisotropy.}
    \begin{tabular}{c|ccccc}
    \toprule
    \textbf{Semantic Groups} & LGA & $\mathbb{C}_{0}$ & $\mathbb{C}_{1}$ & $\mathbb{C}_{2}$ & $\mathbb{C}_{3}$ \\
    \midrule
    \textbf{IoU}(\%) & 44.41 & 45.33 & 45.35 & \textbf{45.53} & 45.26 \\
    \textbf{mIoU}(\%) & 14.16 & 14.69 & 15.59 & \textbf{15.78} & 15.64 \\
    \bottomrule
    \end{tabular}
    \label{tab:ablation-csa-sem}
\end{table}

\subsubsection{Ablation on Cubic Semantic Anisotropy}
In this section, we exploit the impact of different designs for the cubic semantic anisotropy (CSA) module to further investigate its effectiveness. Specifically, we conduct parameter analyses from the following aspects:
\begin{itemize}
    \item the weight parameters $w_e, w_v$ for aggregating reassigned semantic differences
    \item the weight parameters $\alpha, \beta$ for linear projection
\end{itemize}
We equip the baseline method with the MVT and CSA modules as the experimental setting for the following results, in order to keep off the synergetic effect between the CSA and CDA modules. Table~\ref{tab:ablation-csa-weight} presents the ablation results on the weight parameters $w_e, w_v$. In a $3\times 3\times 3$ cubic area, there are total 26 neighbors to the center voxel, which can be classified as 6 surface-adjacent voxels, 12 edge-adjacent voxels, and 8 vertex-adjacent voxels. Considering that the surface-adjacent voxels are directly connected to the center voxel, the weighted number of surface-adjacent voxels should be larger that that of edge-adjacent and corner-adjacent numbers, i.e. $6 > w_e\times 12$ and $6 > w_c\times 8$. Therefore, we range the value of $w_e$ in $\{0.1, 0.3\}$ and $w_c$ in $\{0.1, 0.3, 0.5\}$. As illustrated in Table~\ref{tab:ablation-csa-weight}, the experimental results of different configurations yield similar IoU scores and differ in mIoU performances. This is because the CSA module focuses on the semantic-aware significance of each voxel, leading to more impact on the class-ware semantic scene completion metrics than class-agnostic scene completion metrics. We set $w_e=0.1$ and $w_c=0.3$ to obtain the best mIoU scores. Table~\ref{tab:ablation-csa-proj} demonstrates the ablation experiments on the projection parameters $\alpha, \beta$, where we set $\alpha=1.0$ to keep the original loss and range the value of $\beta$ in $\{0.1, 0.3, 0.5, 0.7, 0.9\}$ to scale the weight of reassigned semantic differences. The best performance is achieved with $\beta=0.5$.

Furthermore, we also evaluate the impact of different groups for the semantic reassignment block. In concrete, we conduct ablation experiments with four configurations for the semantic group:
\begin{itemize}
    \item $\mathbb{C}_{0}$: \{``\textit{empty}", ``\textit{occupied}"\}
    \item $\mathbb{C}_{1}$: \{``\textit{empty}", ``\textit{foreground}", ``\textit{background}"\}
    \item $\mathbb{C}_{2}$: \{``\textit{empty}", ``\textit{vehicle}", ``\textit{human}", ``\textit{ground}", ``\textit{building}", ``\textit{infrastructure}", ``\textit{plant}"\}
    \item $\mathbb{C}_{3}$: initial semantic classes without grouping
\end{itemize}
As illustrated in Table~\ref{tab:ablation-csa-sem}, the best performance is achieved with configuration $\mathbb{C}_{2}$, where similar semantic classes are appropriately clustered. The group settings $\mathbb{C}_{0}$ and $\mathbb{C}_{1}$ are too coarse to correctly identify class-aware semantic significance of each voxel, leading to inferior mIoU scores. On the other hand, semantic group $\mathbb{C}_{3}$ treats each semantic class equally but ignores the different correlations among the semantic classes, suffering from unnecessary semantic differences between similar semantic classes.
Furthermore, our CSA with different group configurations achieves consistent performance improvements over LGA, demonstrating its effectiveness in identifying contextual semantic significance with a comprehensive consideration of cubic neighborhood.

\begin{table}[t]
    \centering
    \caption{Ablation study on the SemanticKITTI validation set for critical voxel number in critical distribution alignment.}
    \begin{tabular}{c|ccccc}
    \toprule
    \textbf{Critical Voxels} $k$ & 0 & 1024 & 2048 & 4096 & 8192 \\
    \midrule
    \textbf{IoU}(\%) & 44.74 & 45.36 & 45.60 & \textbf{45.68} & 45.66 \\
    \textbf{mIoU}(\%) & 14.28 & 15.12 & 15.38 & \textbf{15.87} & 15.72 \\
    \bottomrule
    \end{tabular}
    \label{tab:ablation-cda}
\end{table}

\begin{table}[t]
    \centering
    \caption{Ablation study on the SemanticKITTI validation set for different distance ranges of selected critical voxels in critical distribution alignment.}
    \begin{tabular}{c|ccc}
    \toprule
    \textbf{Distance Ranges} & $0\sim12.8m$ & $0\sim25.6m$ & $0\sim51.2m$ \\
    \midrule
    \textbf{IoU}(\%) & 45.05 & 45.21 & \textbf{45.68} \\
    \textbf{mIoU}(\%) & 14.89 & 15.33 & \textbf{15.87} \\
    \bottomrule
    \end{tabular}
    \label{tab:ablation-cda-range}
\end{table}

\begin{table}[t]
    \centering
    \caption{Ablation study on the SemanticKITTI validation set for the weight parameter of the circulated loss in the final training loss.}
    \begin{tabular}{c|ccccc}
    \toprule
    \textbf{Circ Weight} $\lambda$ & 0.3 & 0.5 & 0.7 & 1.0 & 3.0 \\
    \midrule
    \textbf{IoU}(\%) & 46.82 & 46.77 & 46.81 & \textbf{46.93} & 46.68 \\
    \textbf{mIoU}(\%) & 15.84 & 16.15 & 16.26 & \textbf{16.74} & 16.36 \\
    \bottomrule
    \end{tabular}
    \label{tab:ablation-circ}
\end{table}

\begin{table}[t]
    \centering
    \caption{Efficiency evaluation on the SemanticKITTI validation set against the best comparison method SGN.}
    \resizebox{1.0\linewidth}{!}{
    \begin{tabular}{c|cccc}
    \toprule
    \textbf{Methods} & Params & Memory & GFLOPs & Inference \\
    \midrule
    SGN & 28.16M & 15.83G & 725.05 & 0.71s/img \\
    MRA(k=2048) & 32.33M & 15.03G & 877.63 & 0.89s/img \\
    MRA(k=4096) & 32.33M & 15.03G & 877.63 & 0.90s/img \\
    MRA(k=8192) & 32.33M & 15.04G & 877.63 & 0.90s/img \\
    \bottomrule
    \end{tabular}
    }
    \label{tab:efficiency}
\end{table}

\subsubsection{Ablation on Critical Distribution Alignment}
To get further in-dpeth understanding on the effectiveness of the critical distribution alignment (CDA) module, we conduct ablation experiments with different number of selected critical voxels $k$. To avoid the impact of the synergy between the CDA and CSA module, we only integrate the MVT and CDA modules to the baseline and range the value of $k$ in $\{0, 1024, 2048, 4096, 8192\}$, where setting $k=0$ indicates only employing the MVT module. The results are shown in Table~\ref{tab:ablation-cda}. It can be observed that different values of $k$ achieve performance improvements on both IoU and mIoU scores consistently. As the number of selected values gradually increase from 1024 to 4096, the overall performance keeps boosting due to more comprehensive exploitation of discriminative semantics. But the model performance slightly decreases as the value of $k$ increases form 4096 to 8192.
Selection 8192 critical voxels would include a large proportion of low-resolution features ($64\times64\times8$), inevitably introducing noise from non-informative regions and disturbing distribution alignment from truly discriminative areas. Empirically, selecting 4096 voxels provides a balance where semantically significant regions are sufficiently covered, while the proportion of non-informative voxels remains controlled, achieving optimal performance.
Table~\ref{tab:ablation-cda-range} presents the ablation results on varying $k$ across different distance ranges, with critical voxels constrained in $0\sim12.8m, 0\sim25.6m, 0\sim51.2m$, from near to far, respectively. It can be observed that SSC performance improves as the distance range extends, demonstrating the effectiveness of critical distribution alignment between different resolutions in promoting feature consistency and compensating for sparse voxel supervision, especially in distant regions with sparse and confusing semantics.

\subsubsection{Parameter Analysis on Training Loss}
In Table~\ref{tab:ablation-circ}, we present the experimental results on the weight parameter $\lambda$ of the Circulated Loss $L_{\rm circ}$ in the final training loss, and $\lambda=1.0$ is set for the best performance.
The critical distribution alignment enforces semantic distribution consistency of critical voxel features across resolutions, and achieves consistent performance improvements. Without critical distribution alignment, the model lacks explicit guidance to maintain consistent feature distributions around critical regions, leading to inconsistent contextual semantics for performance degradation.

\subsection{Efficiency Evaluation}
We evaluate the computational complexity and inference efficiency of MRA compared to the best-performing baseline SGN on the SemanticKITTI validation set. As shown in Table~\ref{tab:efficiency}, MRA introduces a moderate increase in parameter count (32.33M vs. 28.16M), GFLOPs (877.63 vs. 725.05) and inference time (0.90s vs 0.71s), primarily due to the multi-resolution design. Despite this, MRA achieves comparable GPU memory usage (15.03G vs. 15.83G), benefiting from efficient sparse voxel selection and alignment operations. The multi-resolution design introduces additional computation, but remains within a reasonable overhead considering the significant performance improvements achieved. We further analyze the impact of varying the number of selected critical voxels ($k$) on model efficiency. Different values of $k$ slightly affect inference time with negligible effect on GPU memory and computation load, indicating the critical voxel selection strategy in MRA is computationally efficient.

\subsection{Generalization on nuScenes Dataset}
Table~\ref{tab:nuscenes} provides the comparison results on the Surround-nuScenes~\cite{wei2023surroundocc} dataset to demonstrate the generalization ability of our MRA approach. It can be observed that MRA achieves consistent performance improvements of \textbf{1.51\%} IoU and \textbf{1.14\%} mIoU, showcasing its robustness across different datasets. It is worth noting that SurroundOcc~\cite{wei2023surroundocc} also employs multi-resolution representations, but the multi-resolution representations are simply utilized to provide multi-scale supervision signals without further alignment. In contrast, our MRA focuses on the alignment across multi-resolution representations to promote semantic consistency, leading to improved SSC performance.

\begin{table}[t]
    \centering
    \caption{Semantic scene completion results on the Surround-nuScenes~\cite{wei2023surroundocc} dataset. Best results are highlighted in \textbf{bold}, and second-best results are \underline{underlined}.}
    \begin{tabular}{l|c|c}
    \toprule
    \textbf{Methods} & \textbf{IoU} & \textbf{mIoU} \\
    \midrule
    \midrule
    MonoScene\cite{cao2022monoscene} & 23.96 & 7.31  \\
    Atlas\cite{murez2020atlas} & 28.66 & 15.00 \\
    BEVFormer\cite{li2022bevformer} & 30.50 & 16.75 \\
    TPVFormer\cite{huang2023tri} & 30.86 & 17.10 \\
    SurroundOcc\cite{wei2023surroundocc} & \underline{31.49} & \underline{20.30} \\
    \rowcolor{gray!10} 
    \textbf{MRA(Ours)} & \textbf{33.00} & \textbf{21.44} \\
    \bottomrule
    \end{tabular}
    \label{tab:nuscenes}
\end{table}

\begin{figure}[t]
\centering
\includegraphics[width=1.0\columnwidth]{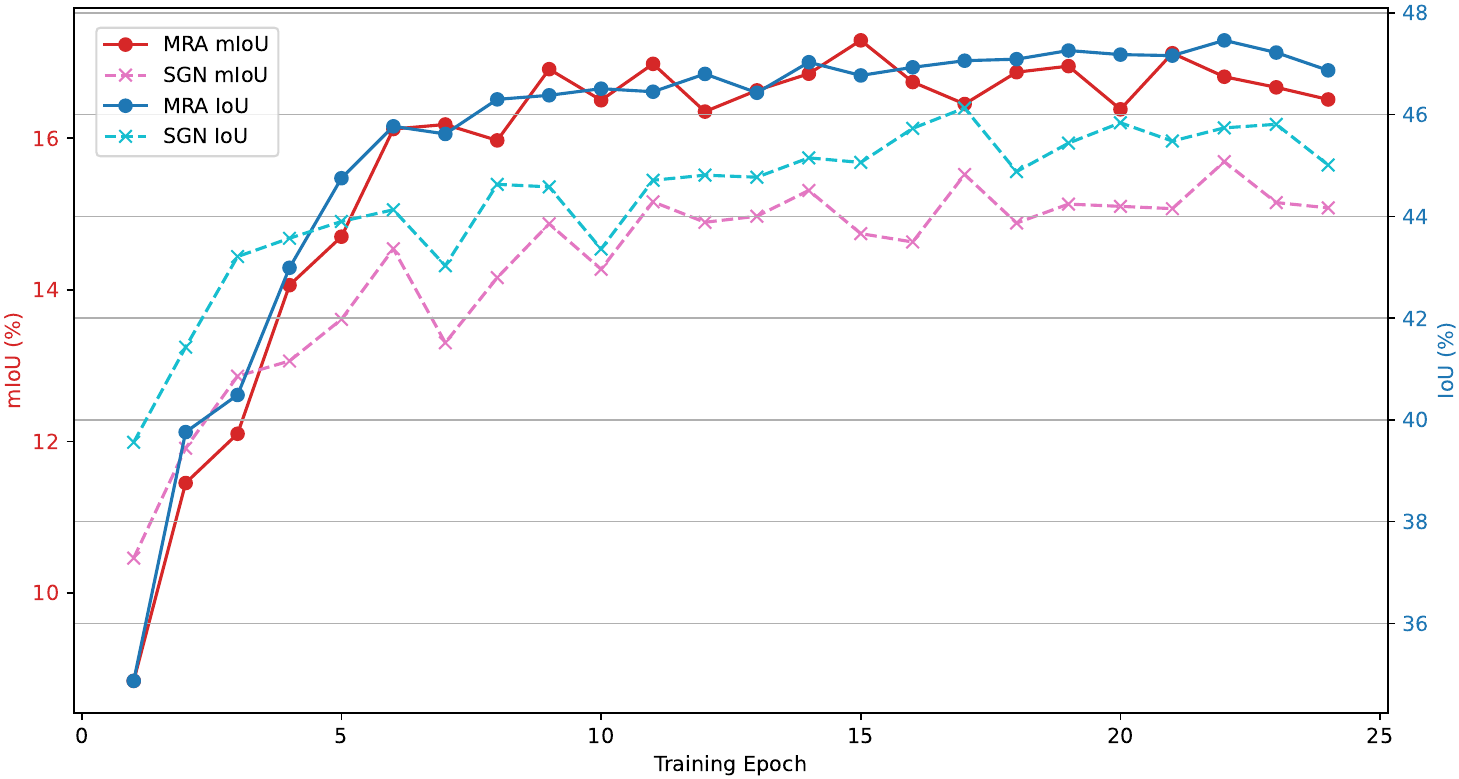}
\caption{Training convergence curves of mIoU and IoU on the SemanticKITTI validation set, comparing MRA and SGN.}
\label{converge}
\end{figure}

\subsection{Training Convergence}
Figure~\ref{converge} presents the training convergence curves concerning mIoU and IoU over 24 epochs, comparing our MAR approach against the best comparison method SGN. It can be observed that MRA exhibits faster convergence in the early training stages, achieving higher mIoU and IoU scores than SGN within the first 10 epochs. Additionally, MRA also maintains a more stable convergence trajectory with smaller performance fluctuations in the later training stages, further demonstrating its effectiveness in resolving voxel sparsity with improved optimization process.

\section{Conclusion}
\label{sec:conclusion}
In this paper, we address the challenge of voxel sparsity in the 3D semantic scene completion (SSC) task and propose the Multi-Resolution Alignment (MRA) approach, which improves the SSC performance by exploiting the self-consistency across multi-resolution 3D features as the complementary supervision signals to the original supervision from sparse voxel labels. The Multi-resolution View Transformer (MVT) module is introduced to project 2D image features into multi-resolution 3D features, facilitating seed feature alignment and enabling the propagation of discriminate semantics over the whole scene. The Cubic Semantic Anisotropy (CSA) module further identifies the semantic-aware significance of each voxel, which executes semantic reassignment to cluster similar categories and aggregates semantic differences with comprehensive consideration of surface-, edge-, and vertex-adjacent voxels in the cubic neighborhood. The Critical Distribution Alignment (CDA) module selects critical voxels with the guidance of CSA and employs the circulated loss as auxiliary supervision on the critical distribution consistency. The proposed MRA approach achieves state-of-the-art camera-based 3D semantic scene completion performance on the SemanticKITTI and SSCBench-KITTI-360 benchmarks, showcasing the effectiveness on addressing the challenge of voxel sparsity. 
While MRA introduces a moderate increase in computational resources, this overhead is considered a reasonable trade-off given the consistent and significant performance improvements achieved.
We hope the proposed MRA approach can promote the exploration of consistent SSC models with self-alignment for improved scene understanding with sparse voxel labels.

Despite MRA's effectiveness, we have also detected failure cases in distant areas with small-scale objects and occlusion issues, where low-quality semantic features lead to incomplete and false semantic scene completion results.
In the future, we aim to incorporate 3D geometric information, including structural priors and motion patterns of target objects, to enhance accurate representation in distant areas and modeling of scene dynamics, facilitating 4D scene understanding and generation for broader applications in real-world autonomous driving scenarios.

\bibliographystyle{IEEEtran}
\bibliography{main}

\end{document}